\documentclass{hld2025} % Include author names
\title[Refresh-Scaling the Memory of Balanced Adam]{Refresh-Scaling the Memory of Balanced Adam}

\hldauthor{%
\Name{Alberto Fernández-Hernández} \Email{a.fernandez@upv.es}\\
\addr Universitat Politècnica de València, València, Spain
\AND
\Name{Cristian Pérez-Corral} \Email{cpercor@upv.es}\\
\addr Universitat Politècnica de València, València, Spain
\AND
\Name{Jose I. Mestre} \Email{jimesmir@upv.es}\\
\addr Universitat Politècnica de València, València, Spain
\AND
\Name{Manuel F. Dolz} \Email{dolzm@uji.es}\\
\addr Universitat Jaume I, Castelló de la Plana, Spain
\AND
\Name{Enrique S. Quintana-Ortí} \Email{quintana@upv.es}\\
\addr Universitat Politècnica de València, València, Spain
}

\usepackage{booktabs}
\usepackage{graphicx}
\usepackage{adjustbox}
\usepackage{float}
\usepackage{placeins}
\usepackage{microtype}
\begin{document}

\maketitle

\begin{abstract}%
Recent evidence suggests that Adam performs robustly when its momentum parameters are tied, $\beta_1=\beta_2$, reducing the optimizer to a single remaining parameter. However, how this parameter should be set remains poorly understood. We argue that, in balanced Adam, $\beta$ should not be treated as a dimensionless constant: it defines a statistical memory horizon $H_\beta=(1-\beta)^{-1}$. In terms of the effective learning horizon $T_{\mathrm{ES}}$, estimated from the validation trajectory, we study the refresh count $R_\beta=(1-\beta)T_{\mathrm{ES}}$, which measures how many times Adam renews its internal statistics during the useful phase of training. Across 11 vision and language experiments, we find that choosing $\beta$ so that $R_\beta\approx1000$ selects different $\beta$ values depending on the training scale, yet improves robustness over the best fixed-beta baseline. Compared with the strongest fixed choice $\beta=0.944$, the refresh rule improves worst-case robustness, reducing the maximum relative gap in validation loss by 33.4\%, while bringing all 11 runs within 1\% of their validation oracle. These results suggest that the remaining hyperparameter of balanced Adam is more naturally viewed as a memory-scale variable than as a fixed constant. This provides a simple budget-aware perspective on optimizer scaling and opens a path toward treating Adam's momentum as part of the learning dynamics rather than as a static default.
\end{abstract}

%\begin{keywords}%
%  List of keywords%
%\end{keywords}

% TL; DR: Once Adam is balanced, the remaining $\beta$ controls a memory clock. That clock should be scaled with the effective learning horizon, and empirically a refresh count around 10^3 is a robust operating point.

\section{Introduction}

Adam \citep{kingma2015adam} remains the default optimizer for modern deep learning, yet one of its most basic design choices is still poorly understood: the use of two different momentum parameters, \(\beta_1\) and \(\beta_2\). The standard setting \((\beta_1,\beta_2)=(0.9,0.999)\), or variants with a smaller second-moment coefficient such as \(\beta_2=0.95\) in large language model training \citep{touvron2023llama}, assigns different time scales to the first and second moment estimates. However, recent empirical evidence suggests that this separation may not be necessary. In particular, large-scale studies of language model training \citep{orvieto2025secret} have shown that tying the two parameters, \(\beta_1=\beta_2\), preserves near-optimal performance while turning Adam into a simpler one-parameter optimizer.

These results shift the question. If Adam is used in the balanced regime ($\beta_1=\beta_2=\beta$), then the remaining problem is no longer how to tune two independent momentum coefficients, but how to choose the single memory scale of the optimizer. Treating \(\beta\) as a fixed dimensionless constant is unsatisfactory: the same value of \(\beta\) represents a very different dynamical regime in a short training run from that in a long one.

This suggests that the remaining parameter of balanced Adam should be studied as a scaling variable. Rather than asking for a universal value of \(\beta\), we ask how Adam's statistical memory should scale with the effective duration of learning. To this end, we define an effective learning horizon \(T_{\mathrm{ES}}\), a validation-based estimate of the training time over which learning remains useful, and introduce the refresh count $R_\beta = (1-\beta)T_{\mathrm{ES}}.$ This quantity measures how many times Adam refreshes its internal statistics during the useful phase of training.

Our main empirical finding is that balanced Adam behaves robustly when $\beta$ is chosen so that this refresh count $R_\beta$ is kept constant. Across a heterogeneous suite of vision and language experiments, the simple rule
\[
    R_\beta \approx 1000,
    \qquad
    \beta = 1 - \frac{1000}{T_{\mathrm{ES}}},
\]
selects different values of \(\beta\) depending on the effective training horizon, but yields a concentrated refresh scale. Compared with the best fixed-\(\beta\) baseline in our grid, \(\beta=0.94377\), this rule improves train robustness and validation tail risk while preserving time-to-target performance.

The contribution of this work is a scale-aware interpretation of Adam's momentum parameter. Once Adam is properly balanced, $\beta$ emerges as a principled control of the optimizer's statistical memory clock, which can be matched to the effective learning horizon.

\section{Balanced Adam as a One-Clock Optimizer}

The original formulation of Adam, presented by  \citet{kingma2015adam}, maintains, at each optimization step $t$, exponential moving averages of the gradient and squared gradient,
\[
    m_t = \beta_1 m_{t-1} + (1-\beta_1)g_t,
    \qquad
    v_t = \beta_2 v_{t-1} + (1-\beta_2)g_t^2,
\]
and updates parameters using the normalized direction \(m_t/\sqrt{v_t}\), up to bias correction and numerical stabilization. In the standard formulation, \(m_t\) and \(v_t\) evolve on different time scales. This gives Adam two independent memory horizons, one for the gradient mean $m_t$ and one for the gradient magnitude $v_t$.

Several recent works indicate that this two-clock view may be unnecessarily complex. Empirically, \citet{orvieto2025secret} found that the constraint \(\beta_1=\beta_2\) retains strong performance across extensive language model training experiments, while also enabling a statistical interpretation of Adam as an online estimator of gradient mean and variance. Dynamically, \citet{fernandez2026balancedadam} showed that the same balanced regime is singled out by a gradient-scale invariance principle: Adam cancels its first-order sensitivity to multiplicative changes in gradient scale if and only if the two momentum parameters are tied. Earlier studies of Adam's dynamics, conducted by \citet{ma2022qualitative}, also observed that making the two momentum factors close can stabilize training and reduce oscillatory or spiking behavior.

In this paper we take the balanced regime as the starting point. Setting $\beta_1=\beta_2=\beta$ turns Adam into a one-clock optimizer. Both internal statistics now evolve on the same exponential time scale, whose effective horizon is $H_\beta = {1}/{(1-\beta)}.$ This quantity has a direct dynamical meaning: it is the approximate number of recent optimization steps that dominate Adam's internal estimates.

However, a memory horizon measured in raw steps is not by itself scale-aware. A horizon of \(H_\beta=1000\) may be short in a million-step run, but extremely long in a ten-thousand-step run. We therefore compare the memory horizon to the effective learning horizon \(T_{\mathrm{ES}}\), obtaining the refresh count $R_\beta = {T_{\mathrm{ES}}}/{H_\beta} = (1-\beta)T_{\mathrm{ES}}.$ 

This quantity measures how many times Adam refreshes its internal statistics during the useful phase of learning. Our hypothesis is that the remaining parameter of balanced Adam should be chosen by keeping this refresh count $R_\beta$ approximately fixed, rather than by fixing \(\beta\) itself:
\[
    R_\beta \approx R_0,
    \qquad
    \beta = 1-\frac{R_0}{T_{\mathrm{ES}}}.
\]
The rest of the paper tests this refresh-scaling hypothesis by identifying a suitable value of \(R_0\) and comparing the resulting rule for choosing \(\beta\) against the baseline of using a fixed \(\beta\).

\section{Calibrating Adam's Refresh Scale}

We evaluate the refresh-scaling hypothesis on 11 balanced Adam \citep{kingma2015adam} or AdamW \citep{loshchilov2019decoupled} experiments spanning vision and language models. The suite includes Llama60M \citep{touvron2023llama}, NanoGPT \citep{karpathy2022nanogpt}, ResNet50 \citep{he2016resnet}, ViT-B/16 \citep{dosovitskiy2021vit}, EfficientNet-B0 \citep{tan2019efficientnet}, Swin-T \citep{liu2021swin}, and T5-small \citep{raffel2020t5}, trained on C4 \citep{raffel2020t5}, SlimPajama \citep{shen2023slimpajama}, OpenWebText \citep{gokaslan2019openwebtext}, WikiText-103 \citep{merity2017wikitext}, Food-101 \citep{bossard2014food101}, ImageNet100 \citep{deng2009imagenet}, CIFAR-100 \citep{krizhevsky2009cifar}, TinyImageNet \citep{le2015tinyimagenet}, Stanford Cars \citep{krause2013cars}, Caltech-256 \citep{griffin2007caltech256}, and BookCorpus \citep{zhu2015bookcorpus}. Full training details, including optimizer variant, batch size, scheduler, weight decay, seeds, and evaluation frequency, are reported in Appendix~\ref{app:experimental-details}.
All runs impose the diagonal constraint $ \beta_1=\beta_2=\beta.$ For each experiment, we sweep a common grid of 13 $\beta$ values, ranging from \(0\) to \(0.999\), with denser spacing near \(1\). Each configuration is first evaluated with a single random seed. We then select the five configurations with the lowest validation loss and rerun each of them with two additional seeds, allowing the most promising settings to be assessed more robustly. The $\beta$-sweep curves are shown in Appendix~\ref{app:beta-sweeps}. They exhibit the expected near-convex shape when plotted on a logarithmic scale in \(1-\beta\), with a clear task-dependent optimum. 

For each experiment \(p\), the relative gap of a $\beta$ value is defined as
\[
\mathrm{gap}_p(\beta)
=
\frac{
    L_p(\beta)-L_p(\beta^\star)
}{
    L_p(\beta^\star)
},
\qquad
\beta^\star
=
\arg\min_{\beta\in \mathcal{B}} L_p(\beta).
\]
Here, \(L_p(\beta)\) denotes the minimum validation loss achieved in experiment \(p\) when trained with $\beta$ value \(\beta\), and \(\mathcal{B}\) is the swept $\beta$ grid.

The effective horizon \(T_{\mathrm{ES}}\) is estimated from the early-stopping times of the two best $\beta$ values in each experiment. We compute both early-stopping times using patience equal to \(10\%\) of the total training budget and \(\texttt{min\_delta=0}\), average them, and round the result to one significant digit. This deliberately coarse estimate reflects the intended use of the rule: \(\beta\) should be inferred from an approximate initial estimate of the number of useful training iterations, not from a finely tuned stopping time. A more detailed analysis of \(T_{\mathrm{ES}}\) and the robustness of the criterion to this choice is provided in Appendix~\ref{app:time-to-target}. 

We split the 11 experiments into 8 development experiments (from Experiment 1 to 8) and 3 held-out validation experiments (from Experiment 9 to 11). The development set is used to calibrate the refresh target \(R_0\), while the held-out set is used only for evaluation. We then sweep rounded refresh targets $ R_0 \in \{300,200,\ldots,2000\}. $ For each \(R_0\), we compute $ \beta_{\mathrm{ref}}(T_{\mathrm{ES}})=1-{R_0}/{T_{\mathrm{ES}}}, $ project this value to the closest $\beta$ in the grid, and evaluate the resulting relative gaps on the development experiments. Among the rounded candidates, we select \(R_0=1000\) because it minimizes the maximum relative gap on the development set. This choice is not unique at the level of the induced decision rule: several nearby values of \(R_0\) project to the same $\beta$ choices on the discrete grid. Further details on the refresh-rule selection are provided in Appendix~\ref{app:refresh-rule-selection}. We therefore use $ R_0=1000 $ as our refresh rule.

Table~\ref{tab:selected-betas} reports, for each experiment, the estimated effective stopping time \(T_{\mathrm{ES}}\), the oracle $\beta^\star$ within the explored grid in terms of validation loss, the $\beta$ selected by the refresh rule with \(R_0=1000\), as well as the resulting relative gap. The rule selects larger $\beta$ values as the effective horizon increases, with selected values of $\beta$ ranging from \(0.822\) to \(0.968\).

\begin{table}[ht]
\centering
\small
\caption{$\beta$ values selected by the refresh rule \(R_\beta=(1-\beta)T_{\mathrm{ES}}\approx1000\). The first block contains the development experiments used to calibrate \(R_0\), and the second block contains held-out validation experiments; relative gaps are validation-loss gaps relative to the oracle \(\beta^\star\).}
\label{tab:selected-betas}
\setlength{\tabcolsep}{4pt}
\begin{tabular*}{\textwidth}{@{\extracolsep{\fill}}cllcccc}
\toprule
Exp. & Model & Dataset & \(T_{\mathrm{ES}}\) & \(\beta^\star\) & Selected \(\beta\) & Relative gap (\%)\\
\midrule
1 & NanoGPT & WikiText-103 & 6000 & 0.944 & 0.822 & 0.515 \\
2 & NanoGPT & OpenWebText & 10000 & 0.944 & 0.900 & 0.205 \\
3 & Llama60M & C4 & 10000 & 0.944 & 0.900 & 0.202 \\
4 & Llama60M & SlimPajama & 10000 & 0.944 & 0.900 & 0.299 \\
5 & ViT-B/16 & CIFAR-100 & 20000 & 0.944 & 0.944 & 0.000 \\
6 & ViT-B/16 & TinyImageNet & 30000 & 0.944 & 0.968 & 0.738 \\
7 & ResNet50 & Food-101 & 30000 & 0.990 & 0.968 & 0.885 \\
8 & ResNet50 & ImageNet100 & 40000 & 0.994 & 0.968 & 0.408 \\
\midrule
9 & EfficientNet-B0 & Cars & 10000 & 0.900 & 0.900 & 0.000 \\
10 & T5-small & BookCorpus & 10000 & 0.982 & 0.900 & 0.124 \\
11 & Swin-T & Caltech-256 & 20000 & 0.944 & 0.944 & 0.000\,\% \\
\bottomrule
\end{tabular*}
\end{table}

Table~\ref{tab:main-results} summarizes the relative gaps obtained by the fixed-beta baseline $\beta= 0.944$ and by the refresh rule. The table reports results separately on the 8 development experiments used to calibrate \(R_0\), on the 3 held-out validation experiments, and on the full suite of 11 experiments. Aggregate performance is summarized by the mean relative gap (Mean), the maximum relative gap (Max), and Conditional Value at Risk (CVaR) over the worst \(25\%\) of experiments. CVaR is used as a tail-risk metric: it averages the largest relative gaps and therefore measures how badly a rule behaves on its worst cases, rather than only on average.

\begin{table}[ht]
\centering
\small
\caption{Relative gaps to the per-experiment oracle $\beta$. The development block contains the 8 experiments used to calibrate \(R_0\), while the held-out block contains the 3 remaining experiments.}
\label{tab:main-results}
\setlength{\tabcolsep}{5pt}
\begin{tabular*}{\textwidth}{@{\extracolsep{\fill}}lcccccccccc}
\toprule
& \multicolumn{3}{c}{Development} 
& \multicolumn{3}{c}{Held-out} 
& \multicolumn{3}{c}{Global}
& \\
\cmidrule(lr){2-4}
\cmidrule(lr){5-7}
\cmidrule(lr){8-10}
Method 
& Mean & Max & CVaR 
& Mean & Max & CVaR 
& Mean & Max & CVaR 
& Gap \(<1\%\) \\
\midrule
Fixed \(\beta=0.944\)
& \textbf{0.264} & 1.328 & 1.055
& 0.359 & 1.067 & 1.067
& \textbf{0.290} & 1.328 & 1.059
& 9/11 \\

Refresh \(R_0=1000\)
& 0.406 & \textbf{0.885} & \textbf{0.811}
& \textbf{0.041} & \textbf{0.124} & \textbf{0.124}
& 0.307 & \textbf{0.885} & \textbf{0.713}
& \textbf{11/11} \\
\bottomrule
\end{tabular*}
\end{table}

The fixed baseline \(\beta=0.944\) is strong. On the development experiments, it achieves the best mean relative gap, \(0.264\%\), and it also gives the best global mean relative gap, \(0.290\%\). This is expected: as shown in Table~\ref{tab:selected-betas}, the validation oracle is often close to this fixed value. Thus, \(\beta=0.944\) is a competitive global default.

The refresh rule mainly improves robustness rather than average validation loss. Globally, it reduces the maximum relative gap from \(1.328\%\) to \(0.885\%\), a relative reduction of \(33.4\%\), and lowers CVaR from \(1.059\%\) to \(0.713\%\), a relative reduction of \(32.7\%\). It also brings all 11 experiments within \(1\%\) of their validation oracle, compared with 9/11 for the fixed baseline.

\section{Conclusion, Limitations, and Future Work}

Recent evidence suggests that Adam can be simplified by tying its two momentum parameters, \(\beta_1=\beta_2\). This leaves a single open question: \textit{how should the remaining $\beta$ be chosen?} Our results suggest that the natural object is not the raw $\beta$ value, but the statistical refresh count
\[
    R_\beta=(1-\beta)T_{\mathrm{ES}}.
\]
Across 11 vision and language experiments, choosing \(\beta\) so that \(R_\beta=1000\) selects different $\beta$ values depending on the effective learning horizon, while keeping the induced refresh count concentrated around \(10^3\).

The resulting rule is primarily a robustness improvement rather than an average-loss improvement. The fixed baseline remains highly competitive in mean relative gap, but the refresh rule improves the worst-case behavior: the global maximum relative gap decreases from \(1.328\%\) to \(0.885\%\), the global CVaR decreases from \(1.059\%\) to \(0.713\%\), and all 11 experiments fall within \(1\%\) of their validation oracle. This supports the view that a fixed $\beta$ can be a strong practical default, while the refresh rule provides a scale-aware mechanism for choosing $\beta$ as the effective learning horizon changes.

This connects optimizer tuning with the scaling perspective of high-dimensional learning dynamics. The same raw $\beta$ can represent different optimizer regimes at different training horizons. By contrast, the refresh count provides a dimensionless quantity that compares Adam's internal memory to the duration of learning. The empirical evidence presented in this article suggests that balanced Adam operates robustly when its statistical memory is calibrated to provide an approximately stable number of effective refreshes across architectures and datasets.

There are several limitations. First, \(T_{\mathrm{ES}}\) is not known a priori and must be estimated or approximated in practice, so the rule is scale-calibrated rather than fully zero-shot. This estimation may be non-trivial, although the selected \(\beta\) values appear reasonably robust to moderate noise in the scale of \(T_{\mathrm{ES}}\). Second, the experimental suite is deliberately heterogeneous and mixes Adam and AdamW, reflecting practical usage while testing whether the effect persists beyond a narrow optimizer--model--dataset setting. Third, although we evaluate several learning rates, we do not fully re-tune the learning-rate schedule for each \(\beta\); future work could study the interaction between the memory clock and learning rate more systematically, potentially leading to a finer joint rule. Finally, \(R_0=1000\) should not be read as a universal constant, but as an empirical refresh scale that performs consistently across the diverse suite considered here.

Future work should test whether the same refresh scale persists under controlled changes in model size, batch size, token budget, and training length. A particularly important direction is to disentangle architecture-dependent effects: language models, CNNs, and vision transformers may occupy nearby but not identical refresh regimes. More dense logging of validation curves and update statistics would also help estimate \(T_{\mathrm{ES}}\) early in training, enabling its combination with the proposed rule without relying on practitioner experience.

\section*{Acknowledgements}
This research was funded by the projects PID2023-146569NB-C21 and PID2023-146569NB-C22 supported by MICIU/AEI/10.13039/501100011033 and ERDF/UE. Alberto Fernández-Hernández was supported by the predoctoral grant PREP2023-001826 supported by MICIU/AEI/10.13039/501100011033 and ESF+. Cristian Pérez-Corral received support from the \textit{Conselleria de Educación, Cultura, Universidades y Empleo} (reference CIACIF/2024/412) through the European Social Fund Plus 2021–2027 (FSE+) program of the \textit{Comunitat Valenciana}. Jose I. Mestre was supported by the predoctoral grant ACIF/2021/281 of the \emph{Generalitat Valenciana}. Manuel F. Dolz was supported by grant {\small CNS2025-165098} funded by {\small MICIU/AEI/10.13039/501100011033} and by the Plan Gen--T grant {\small CIDEXG/2022/013} of the \emph{Generalitat Valenciana}.

\newpage 

\bibliography{sample}

@inproceedings{kingma2015adam,
  title     = {Adam: A Method for Stochastic Optimization},
  author    = {Kingma, Diederik P. and Ba, Jimmy},
  booktitle = {International Conference on Learning Representations},
  year      = {2015},
  url       = {https://arxiv.org/abs/1412.6980}
}

@misc{orvieto2025secret,
  title         = {In Search of {Adam}'s Secret Sauce},
  author        = {Orvieto, Antonio and Gower, Robert M.},
  year          = {2025},
  eprint        = {2505.21829},
  archivePrefix = {arXiv},
  primaryClass  = {cs.LG},
  url           = {https://arxiv.org/abs/2505.21829}
}

@misc{fernandez2026balancedadam,
  title         = {Why {Adam} Works Better with $\beta_1=\beta_2$: The Missing Gradient Scale Invariance Principle},
  author        = {Fern{\'a}ndez-Hern{\'a}ndez, Alberto and P{\'e}rez-Corral, Cristian and Mestre, Jose I. and Dolz, Manuel F. and Quintana-Ort{\'i}, Enrique S.},
  year          = {2026},
  eprint        = {2601.21739},
  archivePrefix = {arXiv},
  primaryClass  = {cs.LG},
  url           = {https://arxiv.org/abs/2601.21739}
}

@inproceedings{ma2022qualitative,
  title     = {A Qualitative Study of the Dynamic Behavior for Adaptive Gradient Algorithms},
  author    = {Ma, Chao and Wu, Lei and E, Weinan},
  booktitle = {Proceedings of the 2nd Mathematical and Scientific Machine Learning Conference},
  series    = {Proceedings of Machine Learning Research},
  volume    = {145},
  pages     = {671--692},
  year      = {2022},
  publisher = {PMLR},
  url       = {https://proceedings.mlr.press/v145/ma22a.html}
}

@inproceedings{loshchilov2019decoupled,
  title     = {Decoupled Weight Decay Regularization},
  author    = {Loshchilov, Ilya and Hutter, Frank},
  booktitle = {International Conference on Learning Representations},
  year      = {2019},
  url       = {https://openreview.net/forum?id=Bkg6RiCqY7}
}

@misc{touvron2023llama,
  title         = {{LLaMA}: Open and Efficient Foundation Language Models},
  author        = {Touvron, Hugo and Lavril, Thibaut and Izacard, Gautier and Martinet, Xavier and Lachaux, Marie-Anne and Lacroix, Timoth{\'e}e and Rozi{\`e}re, Baptiste and Goyal, Naman and Hambro, Eric and Azhar, Faisal and Rodriguez, Aurelien and Joulin, Armand and Grave, Edouard and Lample, Guillaume},
  year          = {2023},
  eprint        = {2302.13971},
  archivePrefix = {arXiv},
  primaryClass  = {cs.CL},
  url           = {https://arxiv.org/abs/2302.13971}
}

@misc{karpathy2022nanogpt,
  title        = {{nanoGPT}},
  author       = {Karpathy, Andrej},
  year         = {2022},
  howpublished = {\url{https://github.com/karpathy/nanoGPT}},
  note         = {GitHub repository}
}

@inproceedings{he2016resnet,
  title     = {Deep Residual Learning for Image Recognition},
  author    = {He, Kaiming and Zhang, Xiangyu and Ren, Shaoqing and Sun, Jian},
  booktitle = {Proceedings of the IEEE Conference on Computer Vision and Pattern Recognition},
  pages     = {770--778},
  year      = {2016},
  url       = {https://openaccess.thecvf.com/content_cvpr_2016/html/He_Deep_Residual_Learning_CVPR_2016_paper.html}
}

@inproceedings{dosovitskiy2021vit,
  title     = {An Image is Worth 16x16 Words: Transformers for Image Recognition at Scale},
  author    = {Dosovitskiy, Alexey and Beyer, Lucas and Kolesnikov, Alexander and Weissenborn, Dirk and Zhai, Xiaohua and Unterthiner, Thomas and Dehghani, Mostafa and Minderer, Matthias and Heigold, Georg and Gelly, Sylvain and Uszkoreit, Jakob and Houlsby, Neil},
  booktitle = {International Conference on Learning Representations},
  year      = {2021},
  url       = {https://openreview.net/forum?id=YicbFdNTTy}
}

@inproceedings{tan2019efficientnet,
  title     = {EfficientNet: Rethinking Model Scaling for Convolutional Neural Networks},
  author    = {Tan, Mingxing and Le, Quoc V.},
  booktitle = {Proceedings of the 36th International Conference on Machine Learning},
  pages     = {6105--6114},
  year      = {2019},
  publisher = {PMLR},
  url       = {https://proceedings.mlr.press/v97/tan19a.html}
}

@inproceedings{liu2021swin,
  title     = {Swin Transformer: Hierarchical Vision Transformer using Shifted Windows},
  author    = {Liu, Ze and Lin, Yutong and Cao, Yue and Hu, Han and Wei, Yixuan and Zhang, Zheng and Lin, Stephen and Guo, Baining},
  booktitle = {Proceedings of the IEEE/CVF International Conference on Computer Vision},
  pages     = {10012--10022},
  year      = {2021},
  url       = {https://openaccess.thecvf.com/content/ICCV2021/html/Liu_Swin_Transformer_Hierarchical_Vision_Transformer_Using_Shifted_Windows_ICCV_2021_paper.html}
}

@article{raffel2020t5,
  title   = {Exploring the Limits of Transfer Learning with a Unified Text-to-Text Transformer},
  author  = {Raffel, Colin and Shazeer, Noam and Roberts, Adam and Lee, Katherine and Narang, Sharan and Matena, Michael and Zhou, Yanqi and Li, Wei and Liu, Peter J.},
  journal = {Journal of Machine Learning Research},
  volume  = {21},
  number  = {140},
  pages   = {1--67},
  year    = {2020},
  url     = {https://jmlr.org/papers/v21/20-074.html}
}

@misc{shen2023slimpajama,
  title         = {{SlimPajama-DC}: Understanding Data Combinations for {LLM} Training},
  author        = {Shen, Zhiqiang and Tao, Tianhua and Ma, Liqun and Neiswanger, Willie and Liu, Zhengzhong and Wang, Hongyi and Tan, Bowen and Hestness, Joel and Vassilieva, Natalia and Soboleva, Daria and Xing, Eric},
  year          = {2023},
  eprint        = {2309.10818},
  archivePrefix = {arXiv},
  primaryClass  = {cs.CL},
  url           = {https://arxiv.org/abs/2309.10818}
}

@misc{gokaslan2019openwebtext,
  title        = {{OpenWebText} Corpus},
  author       = {Gokaslan, Aaron and Cohen, Vanya and Pavlick, Ellie and Tellex, Stefanie},
  year         = {2019},
  howpublished = {\url{https://skylion007.github.io/OpenWebTextCorpus}},
  note         = {Dataset}
}

@inproceedings{merity2017wikitext,
  title     = {Pointer Sentinel Mixture Models},
  author    = {Merity, Stephen and Xiong, Caiming and Bradbury, James and Socher, Richard},
  booktitle = {International Conference on Learning Representations},
  year      = {2017},
  url       = {https://openreview.net/forum?id=Byj72udxe}
}

@inproceedings{bossard2014food101,
  title     = {{Food-101}: Mining Discriminative Components with Random Forests},
  author    = {Bossard, Lukas and Guillaumin, Matthieu and Van Gool, Luc},
  booktitle = {European Conference on Computer Vision},
  pages     = {446--461},
  year      = {2014},
  publisher = {Springer},
  url       = {https://data.vision.ee.ethz.ch/cvl/datasets_extra/food-101/}
}

@inproceedings{deng2009imagenet,
  title     = {{ImageNet}: A Large-Scale Hierarchical Image Database},
  author    = {Deng, Jia and Dong, Wei and Socher, Richard and Li, Li-Jia and Li, Kai and Fei-Fei, Li},
  booktitle = {2009 IEEE Conference on Computer Vision and Pattern Recognition},
  pages     = {248--255},
  year      = {2009},
  publisher = {IEEE},
  doi       = {10.1109/CVPR.2009.5206848}
}

@techreport{krizhevsky2009cifar,
  title       = {Learning Multiple Layers of Features from Tiny Images},
  author      = {Krizhevsky, Alex and Hinton, Geoffrey},
  institution = {University of Toronto},
  year        = {2009},
  url         = {https://www.cs.toronto.edu/~kriz/learning-features-2009-TR.pdf}
}

@article{le2015tinyimagenet,
  title   = {Tiny ImageNet Visual Recognition Challenge},
  author  = {Le, Ya and Yang, Xuan},
  journal = {CS 231N},
  volume  = {7},
  number  = {7},
  pages   = {3},
  year    = {2015}
}

@inproceedings{krause2013cars,
  title     = {3D Object Representations for Fine-Grained Categorization},
  author    = {Krause, Jonathan and Stark, Michael and Deng, Jia and Fei-Fei, Li},
  booktitle = {Proceedings of the IEEE International Conference on Computer Vision Workshops},
  pages     = {554--561},
  year      = {2013},
  url       = {https://openaccess.thecvf.com/content_iccv_workshops_2013/W19/papers/Krause_3D_Object_Representations_2013_ICCV_paper.pdf}
}

@techreport{griffin2007caltech256,
  title       = {{Caltech-256} Object Category Dataset},
  author      = {Griffin, Gregory and Holub, Alex and Perona, Pietro},
  institution = {California Institute of Technology},
  number      = {CNS-TR-2007-001},
  year        = {2007},
  url         = {https://authors.library.caltech.edu/records/5sv1j-ytw97}
}

@inproceedings{zhu2015bookcorpus,
  title     = {Aligning Books and Movies: Towards Story-Like Visual Explanations by Watching Movies and Reading Books},
  author    = {Zhu, Yukun and Kiros, Ryan and Zemel, Richard and Salakhutdinov, Ruslan and Urtasun, Raquel and Torralba, Antonio and Fidler, Sanja},
  booktitle = {Proceedings of the IEEE International Conference on Computer Vision},
  pages     = {19--27},
  year      = {2015},
  url       = {https://arxiv.org/abs/1506.06724}
}

\appendix
\section{Experimental Details}
\label{app:experimental-details}

\paragraph{General protocol.}
All experiments restrict Adam to the diagonal regime \(\beta_1=\beta_2=\beta\). The common $\beta$ grid is
\[ \mathcal{B} = 
\{0,\ 0.438,\ 0.684,\ 0.822,\ 0.900,\ 0.944,\ 0.968,\ 0.982,\ 0.990,\ 0.994,\ 0.997,\ 0.998,\ 0.999\}.
\]
For every experiment and every $\beta$ in the grid, we first run seed 1. We then select the five best $\beta$ values according to the seed-1 validation sweep and rerun those configurations with seeds 2 and 3 to refine the estimate around the 5 near-optimal values of $\beta$. Therefore, seed-averaged results are available only for these refined $\beta$ windows; outside them, results are based on seed 1. The reported losses are \texttt{train\_loss} and \texttt{val\_loss}; the relative gaps in the paper are computed with respect to the best \texttt{val\_loss} achieved in each experiment. We use bf16 autocast when CUDA/bf16 is available, and fp32 otherwise.

\paragraph{Llama60M on C4.}
We train a Llama model with approximately 60M parameters from the configuration \texttt{llama\_60m.json}, not from a pretrained checkpoint, on English C4 (\texttt{allenai/c4}, configuration \texttt{en}) with causal next-token prediction. We use the \texttt{t5-base} tokenizer, maximum sequence length 256, physical batch size 64, and effective batch size 512 through gradient accumulation of 8, corresponding to 131072 tokens per optimizer update. Training lasts 10000 optimizer steps. The optimizer is Adam, weight decay \(10^{-5}\), maximum learning rate \(10^{-3}\), cosine decay down to \(10^{-4}\), 1000 warmup steps, and no gradient clipping by default. Evaluation is performed on validation every 1000 steps, as well as at the first and final step, using up to \(10^7\) evaluation tokens.

\paragraph{Llama60M on SlimPajama-6B.}
This experiment follows the same protocol as Llama60M on C4: a Llama60M model is trained from scratch with the \texttt{t5-base} tokenizer, sequence length 256, physical batch size 64, effective batch size 512, 10000 optimizer steps, Adam,  weight decay \(10^{-5}\), maximum learning rate \(10^{-3}\), cosine decay to \(10^{-4}\), 1000 warmup steps, and no gradient clipping by default. The dataset is SlimPajama-6B (\texttt{DKYoon/SlimPajama-6B}), loaded from the local cache, using the dataset's training and validation splits. Evaluation is performed every 1000 steps and uses up to \(10^7\) tokens.

\paragraph{NanoGPT on OpenWebText.}
We train from scratch a small GPT-2 style model implemented with \texttt{GPT2LMHeadModel}, using vocabulary size 50304, context length 256, 6 layers, 6 heads, embedding dimension 384, and dropout 0.2 on residual, embedding, and attention modules. The dataset is OpenWebText, pretokenized with GPT-2 BPE and packed into blocks of 256 tokens. We use physical batch size 8 and gradient accumulation 128, yielding 262144 tokens per optimizer update. Training lasts 10000 optimizer steps. The optimizer is AdamW, weight decay 0.01 applied only to parameters with decay, maximum learning rate \(6\cdot10^{-4}\), minimum learning rate \(6\cdot10^{-6}\), 2000 warmup steps, cosine decay, and global gradient clipping 1.0. Evaluation is performed every 500 steps using 50 fixed train and validation batches.

\paragraph{NanoGPT on WikiText-103.}
This experiment uses the same model and training protocol as NanoGPT on OpenWebText: a small GPT-2 style model trained from scratch with context length 256, 6 layers, 6 heads, embedding dimension 384, dropout 0.2, physical batch size 8, gradient accumulation 128, 262144 tokens per update, 10000 optimizer steps, AdamW, weight decay 0.01, maximum learning rate \(6\cdot10^{-4}\), minimum learning rate \(6\cdot10^{-6}\), 2000 warmup steps, cosine decay, and gradient clipping 1.0. The dataset is WikiText-103 raw (\texttt{wikitext-103-raw-v1}), loaded from the local cache, tokenized with GPT-2 BPE, and packed into blocks of 256 tokens. Evaluation is performed every 500 steps using 50 fixed batches.

\paragraph{ResNet50 on Food-101.}
We train ResNet50 from scratch (\texttt{weights=None}) on Food-101. The final head is replaced by dropout with \(p=0.2\) followed by a linear classifier. We use batch size 128, 50 epochs, 591 batches per epoch, and 29550 total steps. The optimizer is AdamW and two parameter groups: weight decay 0.01 for weights and 0 for bias, normalization, and batch-normalization parameters. Although the optimizer is initialized with learning rate \(5\cdot10^{-4}\), the training schedule uses effective base learning rate \(3\cdot10^{-4}\), linear warmup over 5\% of total steps, and cosine decay to \(3\cdot10^{-5}\). The training loss is cross-entropy with label smoothing 0.1; validation uses standard cross-entropy. Augmentations are RandomResizedCrop 224 with scale 0.6--1.0, horizontal flip, ColorJitter 0.2/0.2/0.2/0.1, and standard normalization; validation uses resize 256 followed by center crop 224. Evaluation is performed once per epoch.

\paragraph{ResNet50 on ImageNet100.}
This experiment follows the same protocol as ResNet50 on Food-101, but uses ImageNet100. ResNet50 is trained from scratch with dropout \(p=0.2\) before the final classifier. We use batch size 128, 80 epochs, 1015 batches per epoch, and 81200 total steps. The optimizer is AdamW, weight decay 0.01 for weights and 0 for bias and normalization parameters. The effective base learning rate is \(3\cdot10^{-4}\), with warmup over 5\% of total steps and cosine decay to \(3\cdot10^{-5}\). Training uses cross-entropy with label smoothing 0.1; validation uses standard cross-entropy. Augmentations match the Food-101 setup: RandomResizedCrop 224, horizontal flip, ColorJitter, and normalization; validation uses resize 256 and center crop 224. Evaluation is performed once per epoch.

\paragraph{ViT-B/16 on CIFAR-100.}
We train ViT-B/16 from scratch (\texttt{weights=None}) on CIFAR-100, replacing the head by a linear classifier with 100 output classes. We use batch size 128, 50 epochs, 390 batches per epoch, and 19500 total steps. The optimizer is AdamW, weight decay 0.1 for weights and 0 for bias and normalization parameters. The base learning rate is \(5\cdot10^{-4}\), the minimum learning rate is \(5\cdot10^{-6}\), warmup lasts 1000 steps, and the schedule follows cosine decay. We use global gradient clipping 1.0. The training loss is cross-entropy with label smoothing 0.1; validation uses standard cross-entropy. Augmentations are RandomResizedCrop 224 with scale 0.5--1.0, horizontal flip, RandAugment with 2 operations and magnitude 12, normalization, and RandomErasing with \(p=0.25\); validation uses resize 256 and center crop 224. Evaluation is performed once per epoch.

\paragraph{ViT-B/16 on TinyImageNet.}
This experiment follows the same protocol as ViT-B/16 on CIFAR-100, but uses TinyImageNet. ViT-B/16 is trained from scratch with a linear head for the TinyImageNet classes. We use batch size 128, 50 epochs, 781 batches per epoch, and 39050 total steps. The optimizer is AdamW, weight decay 0.1 for weights, base learning rate \(5\cdot10^{-4}\), minimum learning rate \(5\cdot10^{-6}\), 1000 warmup steps, cosine decay, and global gradient clipping 1.0. Training uses cross-entropy with label smoothing 0.1; validation uses standard cross-entropy. Augmentations match the CIFAR-100 setup: RandomResizedCrop 224, horizontal flip, RandAugment, RandomErasing, and normalization. Evaluation is performed once per epoch.

\paragraph{EfficientNet-B0 on Stanford Cars.}
We train EfficientNet-B0 from scratch (\texttt{weights=None}) on Stanford Cars, replacing the final head by a linear classifier. We use batch size 64, 100 epochs, 128 batches per epoch, and 12800 total steps. The optimizer is Adam with weight decay \(5\cdot10^{-5}\), maximum learning rate \(8\cdot10^{-4}\), minimum learning rate \(10^{-5}\), warmup of \(\max(100,0.05T)\) steps, and cosine decay. We use global gradient clipping 1.0. Training uses cross-entropy with label smoothing 0.05; validation uses standard cross-entropy. Augmentations are RandomResizedCrop 224 with scale 0.7--1.0 and ratio 0.85--1.15, horizontal flip, mild ColorJitter, RandAugment with 1 operation and magnitude 7, normalization, and RandomErasing with \(p=0.1\); validation uses resize 256 and center crop 224. Evaluation is performed once per epoch.

\paragraph{Swin-T on Caltech-256.}
We train Swin-T from scratch (\texttt{weights=None}) on Caltech-256. We construct a split with \(\texttt{split\_seed}=123\), using 60 training images per class and the remaining images for validation. We use batch size 64, 60 epochs, 376 batches per epoch, and 22560 total steps. The optimizer is AdamW, weight decay 0.05 for weights and 0 for bias and normalization parameters. The maximum learning rate is \(4\cdot10^{-4}\), the minimum learning rate is \(4\cdot10^{-6}\), warmup lasts \(\max(100,0.05T)\) steps, and the schedule follows cosine decay. We use global gradient clipping 1.0. Training uses cross-entropy with label smoothing 0.1; validation uses standard cross-entropy. Augmentations are RandomResizedCrop 224 with scale 0.5--1.0, horizontal flip, RandAugment with 2 operations and magnitude 12, normalization, and RandomErasing with \(p=0.25\); validation uses resize 256 and center crop 224. Evaluation is performed once per epoch.

\paragraph{T5-small on BookCorpus.}
We start from the pretrained \texttt{t5-small} checkpoint and continue denoising pretraining on BookCorpus. A validation set of 20000 documents is reserved and the rest is used for training. The objective uses span corruption with \(\texttt{noise\_density}=0.15\), \(\texttt{mean\_noise\_span\_length}=3.0\), maximum input length 256, the \texttt{t5-small} tokenizer, physical batch size 8, effective batch size 64 through gradient accumulation of 8, 10000 optimizer steps, and evaluation batch size 8. The optimizer is AdamW, weight decay 0.01, maximum learning rate \(10^{-4}\), minimum learning rate \(10^{-5}\), warmup ratio 0.1, cosine decay, and gradient clipping 1.0. Evaluation is performed every 1000 optimizer steps on the validation set. The final analysis uses seed 1; additional T5 seeds are treated as dubious and are not used in the main averages.

\paragraph{Code availability.}
All code used to run the $\beta$ sweeps, compute the refresh-rule metrics, and reproduce the tables and appendix analyses is available in the repository
\[
\text{\url{https://github.com/anonymized/Adam_beta_value}}.
\]

\section{Beta Sweep Curves}
\label{app:beta-sweeps}

\begin{figure}[H]
    \centering
    \includegraphics[
        width=\textwidth]{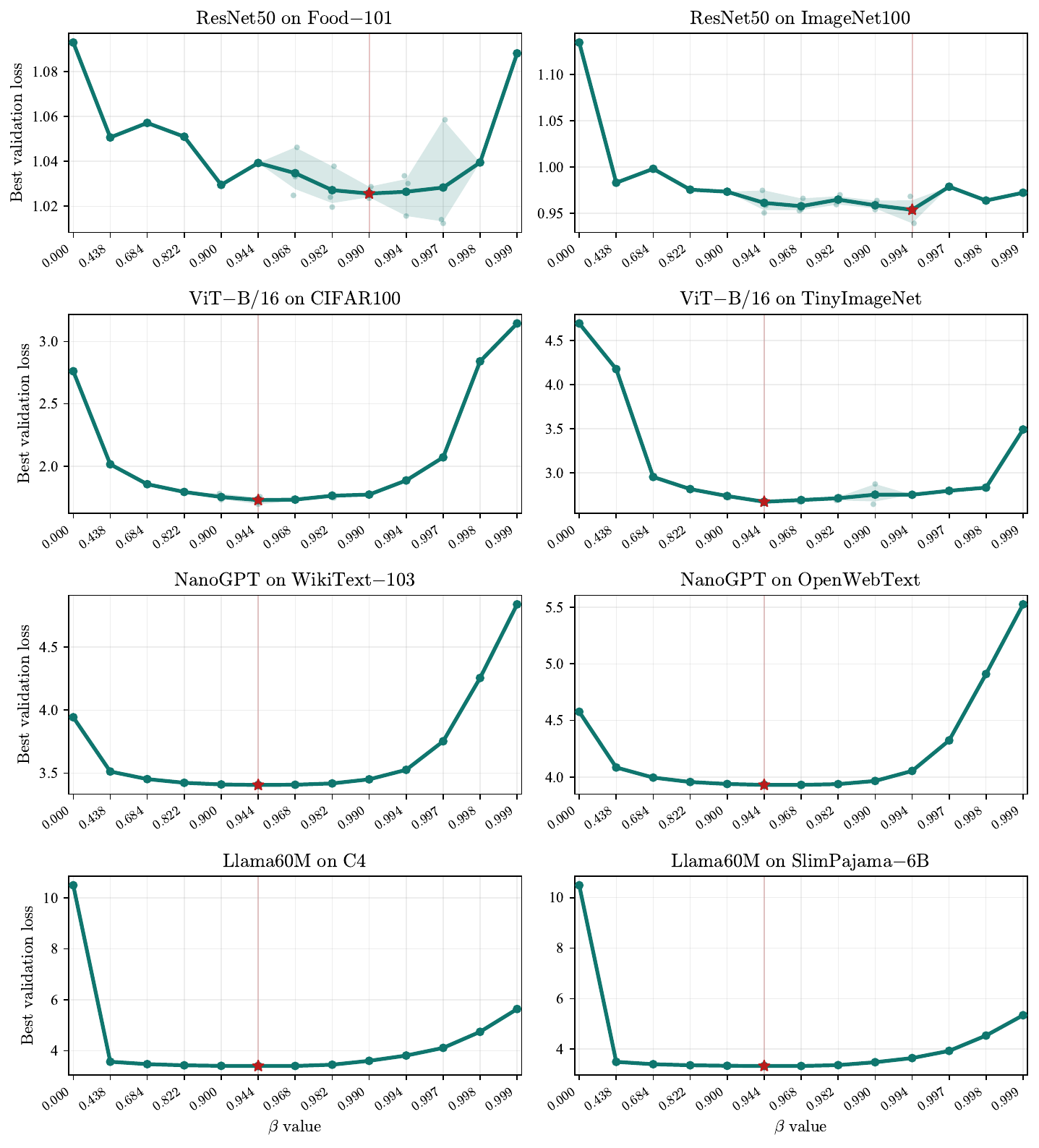}
    \caption{Balanced Adam $\beta$ sweeps for the development experiments.}
    \label{fig:appendix_beta_sweeps_main}
\end{figure}

Figure~\ref{fig:appendix_beta_sweeps_main} shows the balanced-Adam $\beta$ sweeps for the 8 development experiments: ResNet50 on Food-101 and ImageNet100, ViT-B/16 on CIFAR-100 and TinyImageNet, NanoGPT on WikiText-103 and OpenWebText, and Llama60M on C4 and SlimPajama-6B. Each panel reports the best validation loss attained for each value of \(\beta\) on the diagonal \(\beta_1=\beta_2\). When multiple seeds are available, curves use the mean of the seed-level minima and the shaded band reports seed variability. The $\beta$ axis is plotted using the transformation \(u=-\log_{10}(1-\beta)\), with tick labels rendered as $\beta$ values. The vertical marker denotes the $\beta$ minimizing mean validation loss.

\newpage 

\begin{figure}[ht]
    \centering
        \includegraphics[
        width=\textwidth]{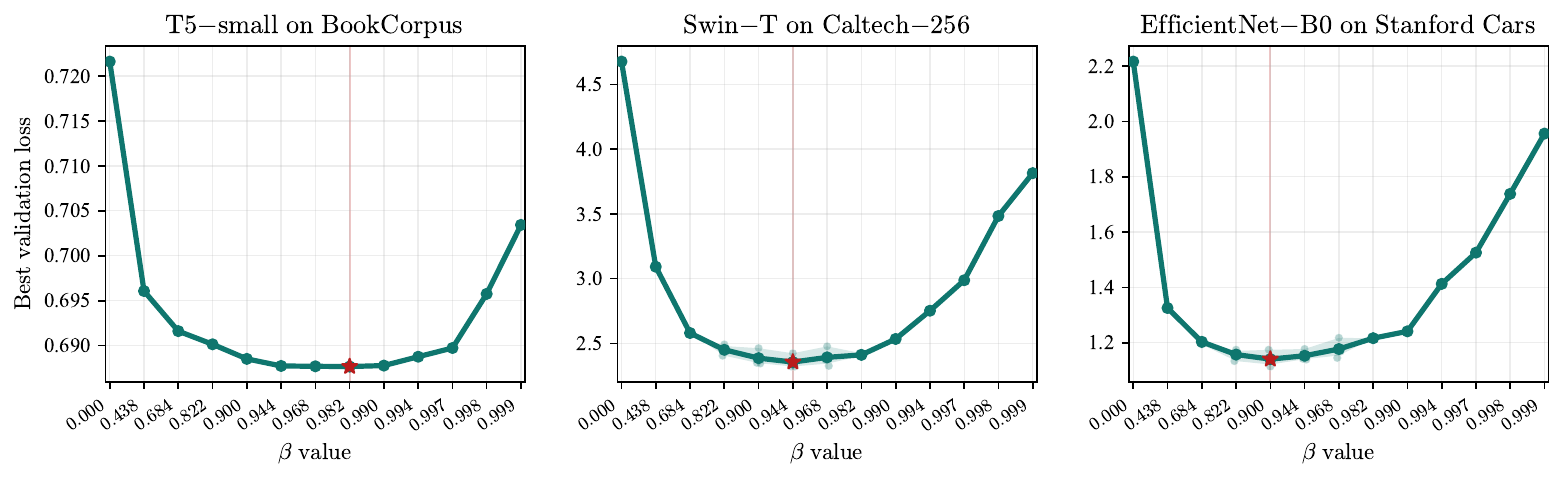}
    \caption{Balanced Adam $\beta$ sweeps for the held-out experiments.}
    \label{fig:appendix_beta_sweeps_validation}
\end{figure}

Figure~\ref{fig:appendix_beta_sweeps_validation} shows the corresponding sweeps for the 3 held-out validation experiments: T5-small on BookCorpus, Swin-T on Caltech-256, and EfficientNet-B0 on Stanford Cars.

\section{Robustness to the Effective Horizon}
\label{app:time-to-target}

The refresh rule requires an estimate of the effective learning horizon \(T_{\mathrm{ES}}\). In the main text, this horizon is used only at a coarse scale: it determines which $\beta$ value is selected after projecting the continuous rule \(\beta=1-R_0/T_{\mathrm{ES}}\) onto the discrete $\beta$ grid. This appendix evaluates how sensitive the method is to errors in this estimate.

\begin{table}[ht]
\centering
\small
\caption{Stability intervals of the $\beta$ selected by the refresh rule \(R_\beta=(1-\beta)T_{\mathrm{ES}}\approx1000\). For each experiment, \(T_{\mathrm{ES}}\) denotes the rounded horizon used by the rule, and the last two columns show the range of \(T_{\mathrm{ES}}\) values for which the selected grid $\beta$ remains unchanged.}
\label{tab:beta-stability-intervals}
\setlength{\tabcolsep}{4pt}
\begin{tabular*}{\textwidth}{@{\extracolsep{\fill}}cllcccc}
\toprule
Exp. & Model & Dataset & selected \(\beta\) & \(T_{\mathrm{ES}}\) & lower \(T_{\mathrm{ES}}\) & upper \(T_{\mathrm{ES}}\) \\
\midrule
1 & NanoGPT & WikiText-103 & 0.822 & 6000 & 4048 & 7199 \\
2 & NanoGPT & OpenWebText & 0.900 & 10000 & 7199 & 12802 \\
3 & Llama60M & C4 & 0.900 & 10000 & 7199 & 12802 \\
4 & Llama60M & SlimPajama & 0.900 & 10000 & 7199 & 12802 \\
5 & ViT-B/16 & CIFAR-100 & 0.944 & 20000 & 12802 & 22766 \\
6 & ViT-B/16 & TinyImageNet & 0.968 & 30000 & 22766 & 40486 \\
7 & ResNet50 & Food-101 & 0.968 & 30000 & 22766 & 40486 \\
8 & ResNet50 & ImageNet100 & 0.968 & 40000 & 22766 & 40486 \\
\midrule
9 & EfficientNet-B0 & Cars & 0.900 & 10000 & 7199 & 12802 \\
10 & T5-small & BookCorpus & 0.900 & 10000 & 7199 & 12802 \\
11 & Swin-T & Caltech-256 & 0.944 & 20000 & 12802 & 22766 \\
\bottomrule
\end{tabular*}
\end{table}

Table~\ref{tab:beta-stability-intervals} reports, for each experiment, the $\beta$ selected by the refresh rule and the range of \(T_{\mathrm{ES}}\) values that would lead to exactly the same selected $\beta$. These intervals are induced by the discretization of the $\beta$ grid: as long as the estimated horizon remains within the reported lower and upper bounds, the selected grid $\beta$ does not change. On average, the relative interval width, measured as \((T_{\mathrm{upper}}-T_{\mathrm{lower}})/T_{\mathrm{lower}}\), is \(77.8\%\). Thus, the selected $\beta$ is stable over a broad range of horizon estimates, often close to a factor of two.

The interval analysis is useful, but conservative. It only checks whether the exact same grid $\beta$ is selected. In practice, if the estimate of \(T_{\mathrm{ES}}\) crosses one of these boundaries, the rule may select a neighboring $\beta$ that still has nearly identical validation loss. We therefore also test the robustness of the method under stochastic perturbations of \(T_{\mathrm{ES}}\).

For each experiment, we replace the estimated horizon by
\[
    T' = T_{\mathrm{ES}}(1+\epsilon),
    \qquad
    \epsilon \sim \mathcal{N}(0,\sigma),
\]
and then apply the same refresh rule using \(T'\). For each \(\sigma>0\), we run 20 perturbations per experiment and report the resulting relative gaps. The first two rows of Table~\ref{tab:refresh-noise-robustness} provide references: the fixed-beta baseline \(\beta=0.944\), and the unperturbed refresh rule with \(\sigma=0\). The remaining rows show increasing levels of multiplicative error in the horizon estimate. For readability, relative gaps in Table~\ref{tab:refresh-noise-robustness} are reported as percentages rather than fractions. The first two rows summarize one value per experiment, hence 11 evaluations. For \(\sigma>0\), each experiment is perturbed 20 times, yielding \(11\times20=220\) evaluations per noise level.

\begin{table}[ht]
\centering
\small
\caption{Robustness of the refresh rule under multiplicative noise in \(T_{\mathrm{ES}}\). For each experiment, we perturb the estimated horizon as \(T'=T_{\mathrm{ES}}(1+\epsilon)\), with \(\epsilon\sim\mathcal{N}(0,\sigma)\), and select the closest grid $\beta$ induced by the refresh rule. For each \(\sigma>0\), results are averaged over 20 perturbations per experiment.}
\label{tab:refresh-noise-robustness}
\setlength{\tabcolsep}{5pt}
\begin{tabular*}{\textwidth}{@{\extracolsep{\fill}}lcccccccccc}
\toprule
& \multicolumn{3}{c}{Development} 
& \multicolumn{3}{c}{Held-out} 
& \multicolumn{3}{c}{Global}
& \\
\cmidrule(lr){2-4}
\cmidrule(lr){5-7}
\cmidrule(lr){8-10}
Method 
& Mean & Max & CVaR 
& Mean & Max & CVaR 
& Mean & Max & CVaR 
& Gap \(<1\%\) \\
\midrule
Fixed \(\beta=0.944\)
& 0.264 & 1.328 & 1.055
& 0.359 & 1.067 & 1.067
& 0.290 & 1.328 & 1.059
& 81.82\% \\

\(\sigma=0\)
& 0.406 & 0.885 & 0.811
& 0.041 & 0.124 & 0.124
& 0.307 & 0.885 & 0.713
& 100.00\% \\

\(\sigma=0.03\)
& 0.430 & 1.156 & 0.864
& 0.041 & 0.124 & 0.124
& 0.324 & 1.156 & 0.789
& 97.73\% \\

\(\sigma=0.06\)
& 0.439 & 1.156 & 0.885
& 0.041 & 0.124 & 0.124
& 0.331 & 1.156 & 0.812
& 96.82\% \\

\(\sigma=0.10\)
& 0.443 & 1.156 & 0.895
& 0.041 & 0.124 & 0.124
& 0.333 & 1.156 & 0.824
& 96.36\% \\

\(\sigma=0.15\)
& 0.454 & 1.156 & 0.926
& 0.184 & 1.544 & 0.644
& 0.380 & 1.544 & 0.946
& 92.27\% \\

\(\sigma=0.20\)
& 0.461 & 1.369 & 0.975
& 0.294 & 1.544 & 1.053
& 0.415 & 1.544 & 1.035
& 89.09\% \\

\(\sigma=0.25\)
& 0.521 & 3.772 & 1.155
& 0.276 & 1.544 & 1.003
& 0.454 & 3.772 & 1.169
& 86.82\% \\
\bottomrule
\end{tabular*}
\end{table}

As expected, performance gradually degrades as the noise level increases. However, the rule remains robust to moderate errors in the estimated horizon. Up to \(\sigma=0.20\), the global CVaR remains below that of the fixed-beta baseline, and the fraction of configurations within \(1\%\) of the oracle remains above the fixed baseline. At \(\sigma=0.25\), the method starts to lose this advantage in global CVaR, although the success rate remains higher than that of the fixed baseline. Overall, the refresh rule does not require a finely tuned estimate of \(T_{\mathrm{ES}}\); an approximate horizon at the correct scale is sufficient for the rule to remain competitive.

\section{Refresh Rule Selection}
\label{app:refresh-rule-selection}

This appendix details how the refresh constant \(R_0\) is selected. The 11 experiments are split into 8 development experiments and 3 held-out validation experiments. The development set contains four model families with two datasets each, and is used to choose \(R_0\). The held-out experiments are kept separate and used only to check whether the selected rule transfers beyond the calibration set.

For each candidate value of \(R_0\), the rule selects a single $\beta$ value through
\(
    \beta_{\mathrm{ref}} = 1-{R_0}/{T_{\mathrm{ES}}},
\)
followed by projection to the closest $\beta$ in the grid. We sweep \(R_0\) from 300 to 2000 in steps of 100 and report the resulting relative gaps in Table~\ref{tab:R0-sweep}. The fixed baseline \(\beta=0.944\) is included as a reference.

\begin{table}[ht] \centering \small \caption{Sweep over the refresh constant \(R_0\). For each value of \(R_0\), the rule selects one $\beta$ per experiment using the current \(T_{\mathrm{ES}}\) estimate, computed as the mean of the two best validation-loss runs and rounded to one significant digit. Relative gaps are validation-loss gaps relative to the per-experiment oracle $\beta$.} \label{tab:R0-sweep} \setlength{\tabcolsep}{4pt} \begin{tabular*}{\textwidth}{@{\extracolsep{\fill}}lcccccccccc} \toprule & \multicolumn{3}{c}{Development} & \multicolumn{3}{c}{Held-out} & \multicolumn{3}{c}{Global} & \\ \cmidrule(lr){2-4} \cmidrule(lr){5-7} \cmidrule(lr){8-10} Method & Mean & Max & CVaR & Mean & Max & CVaR & Mean & Max & CVaR & Gap \(<1\%\) \\ \midrule Fixed \(\beta=0.944\) & \textbf{0.264} & 1.328 & 1.055 & 0.359 & 1.067 & 1.067 & \textbf{0.290} & 1.328 & 1.059 & 9/11 \\ \(R_0=300\) & 0.650 & 3.032 & 2.530 & 1.853 & 3.196 & 3.196 & 0.978 & 3.196 & 2.863 & 7/11 \\ \(R_0=400\) & 0.714 & 3.032 & 2.530 & 1.853 & 3.196 & 3.196 & 1.025 & 3.196 & 2.863 & 7/11 \\ \(R_0=500\) & 0.312 & 1.501 & 1.008 & 0.874 & 1.544 & 1.544 & 0.465 & 1.544 & 1.371 & 8/11 \\ \(R_0=600\) & 0.392 & 1.501 & 1.329 & 0.874 & 1.544 & 1.544 & 0.523 & 1.544 & 1.401 & 7/11 \\ \(R_0=700\) & 0.392 & 1.501 & 1.329 & 0.874 & 1.544 & 1.544 & 0.523 & 1.544 & 1.401 & 7/11 \\ \(R_0=800\) & 0.477 & 1.156 & 1.021 & 0.556 & 1.544 & 1.544 & 0.499 & 1.544 & 1.195 & 9/11 \\ \(R_0=900\) & 0.500 & 1.156 & 1.021 & \textbf{0.041} & \textbf{0.124} & \textbf{0.124} & 0.375 & 1.156 & 0.926 & 10/11 \\ \(R_0=1000\) & 0.406 & \textbf{0.885} & \textbf{0.811} & \textbf{0.041} & \textbf{0.124} & \textbf{0.124} & 0.307 & \textbf{0.885} & \textbf{0.713} & \textbf{11/11} \\ \(R_0=1100\) & 0.406 & \textbf{0.885} & \textbf{0.811} & \textbf{0.041} & \textbf{0.124} & \textbf{0.124} & 0.307 & \textbf{0.885} & \textbf{0.713} & \textbf{11/11} \\ \(R_0=1200\) & 0.406 & \textbf{0.885} & \textbf{0.811} & \textbf{0.041} & \textbf{0.124} & \textbf{0.124} & 0.307 & \textbf{0.885} & \textbf{0.713} & \textbf{11/11} \\ \(R_0=1300\) & 0.406 & \textbf{0.885} & \textbf{0.811} & \textbf{0.041} & \textbf{0.124} & \textbf{0.124} & 0.307 & \textbf{0.885} & \textbf{0.713} & \textbf{11/11} \\ \(R_0=1400\) & 0.580 & 1.328 & 1.119 & 0.602 & 1.446 & 1.446 & 0.586 & 1.446 & 1.228 & 9/11 \\ \(R_0=1500\) & 0.687 & 1.369 & 1.348 & 0.602 & 1.446 & 1.446 & 0.664 & 1.446 & 1.381 & 8/11 \\ \(R_0=1600\) & 0.865 & 1.424 & 1.397 & 1.032 & 1.446 & 1.446 & 0.910 & 1.446 & 1.413 & 6/11 \\ \(R_0=1700\) & 0.865 & 1.424 & 1.397 & 1.032 & 1.446 & 1.446 & 0.910 & 1.446 & 1.413 & 6/11 \\ \(R_0=1800\) & 0.912 & 1.424 & 1.397 & 1.032 & 1.446 & 1.446 & 0.945 & 1.446 & 1.413 & 6/11 \\ \(R_0=1900\) & 0.912 & 1.424 & 1.397 & 1.032 & 1.446 & 1.446 & 0.945 & 1.446 & 1.413 & 6/11 \\ \(R_0=2000\) & 0.912 & 1.424 & 1.397 & 1.032 & 1.446 & 1.446 & 0.945 & 1.446 & 1.413 & 6/11 \\ \bottomrule \end{tabular*} \end{table}

The sweep shows that the method is not sensitive to a finely tuned value of \(R_0\). The fixed baseline has the best mean relative gap on the development set, which is consistent with the main text: \(\beta=0.944\) is a very strong global default. However, its worst-case behavior is less favorable. The range \(R_0\in\{1000, 1100, 1200, 1300\}\) gives the same induced $\beta$ choices on the discrete grid and achieves the best development max relative gap, best development CVaR, best global max relative gap, best global CVaR, and \(11/11\) experiments within \(1\%\) of the oracle.

We therefore choose \(R_0=1000\) as the main refresh constant. This choice is deliberately simple and lies at the beginning of the best-performing plateau. The important point is not that \(1000\) is a sharply optimized constant, but that a broad order-of-magnitude region yields stable and robust $\beta$ choices. This supports the interpretation of the rule as a refresh-scale criterion rather than as an overfitted numerical constant.

\end{document}